\documentclass[sigconf]{acmart}
\usepackage{graphicx}
\usepackage{float}
\usepackage{subfigure}
\usepackage{algpseudocode}
\usepackage{amsmath}

\def\BibTeX{{\rm B\kern-.05em{\sc i\kern-.025em b}\kern-.08emT\kern-.1667em\lower.7ex\hbox{E}\kern-.125emX}}

\copyrightyear{2020}
\acmYear{2020}
\setcopyright{rightsretained}
\acmConference[Multimedia 2020]{ACM Multimedia Grand Challenge}{October 2020}{Seattle, United States}
\acmPrice{15.00}
\acmDOI{10.1145/1122445.1122456}
\acmISBN{978-1-4503-9999-9/18/06}

\author{Hui Lv}
\email{hubrthui@njust.edu.cn}
\affiliation{%
	\institution{Nanjing University of Science and Technology}
	\country{China}
}

\author{Chunyan Xu}
\email{cyx@njust.edu.cn}
\affiliation{%
	\institution{Nanjing University of Science and Technology}
	\country{China}
}
\author{Zhen Cui}
\email{zhen.cui@njust.edu.cn}
\affiliation{%
	\institution{Nanjing University of Science and Technology}
	\country{China}
}
\begin{document}

%
\title{Global Information Guided Video Anomaly Detection}

\renewcommand{\shortauthors}{Lv et al.}

%
\begin{abstract}
Video anomaly detection (VAD) is currently a challenging task due to the complexity of ``anomaly"  as well as the lack of labor-intensive temporal annotations. In this paper, we propose an end-to-end Global Information Guided (GIG) anomaly detection framework for anomaly detection using the video-level annotations (i.e., weak labels). We propose to first mine the global pattern cues by leveraging the weak labels in a GIG module. Then we build a spatial reasoning module to measure the relevance between vectors in spatial domain with the global cue vectors, and select the most related feature vectors for temporal anomaly detection. The experimental results on the CityScene challenge~\cite{CityScene} demonstrate the effectiveness of our model.
\end{abstract}

%
%



%

%

%
\maketitle

\section{Introduction}
Detecting abnormal events in video sequences is a popular task due to the real-world applications such as surveillance and fault detection systems. 
Anomalies are often defined as behavioral or appearance patterns that do not conform to usual patterns~\cite{zhao2011online,hasan2016learning,li2013anomaly}.
And it is a time-and-labor-consuming job to manually identify anomalies in videos.
Therefore, it is a pressing need to develop intelligent computer vision algorithms to analyze the large amount of raw video data and detect video anomalies automatically.
So far the task is extremely hard. The challenges include insufficient annotated data due to the rare occurrences of anomalies, large inter/intra class variations, subjective definition of anomalous events, low resolution of surveillance videos, etc.
Hence, the motion and appearance cues are vital for video anomaly detection. For example, the motion flow of road vehicles can be used to infer accidents. Also appearance information, e.g., fire and smoke, benefits a lot for detecting anomalies like $explosion$. In this paper, we are motivated to explore these motion pattern in temporal domain and appearance information in spatial domain.

The goal of a practical anomaly detection system is to timely signal an activity that deviates normal patterns and identify the time window of the occurring anomaly. 
We propose to leverage the video-level annotations for video anomaly detection under weak supervision. Previously, anomaly detectors tackle the problem in a two-stage manner~\cite{sultani2018real,zhong2019graph,zhu2019motion}.
In the first stage, the authors use well pre-trained models to extract high-level semantic feature vectors from the raw video clips, then they divide each video sequence into segments and group the clip-level feature vectors to segment-level.
In the second stage, they build anomaly detection models by taking segment-level features as input and outputting the corresponding anomaly scores.
The whole process is tedious that one must prepare specific semantic features first and read the saved feature files for anomaly detection. 

\begin{figure*}[t] 
	\centering
	\includegraphics[width=1\textwidth]{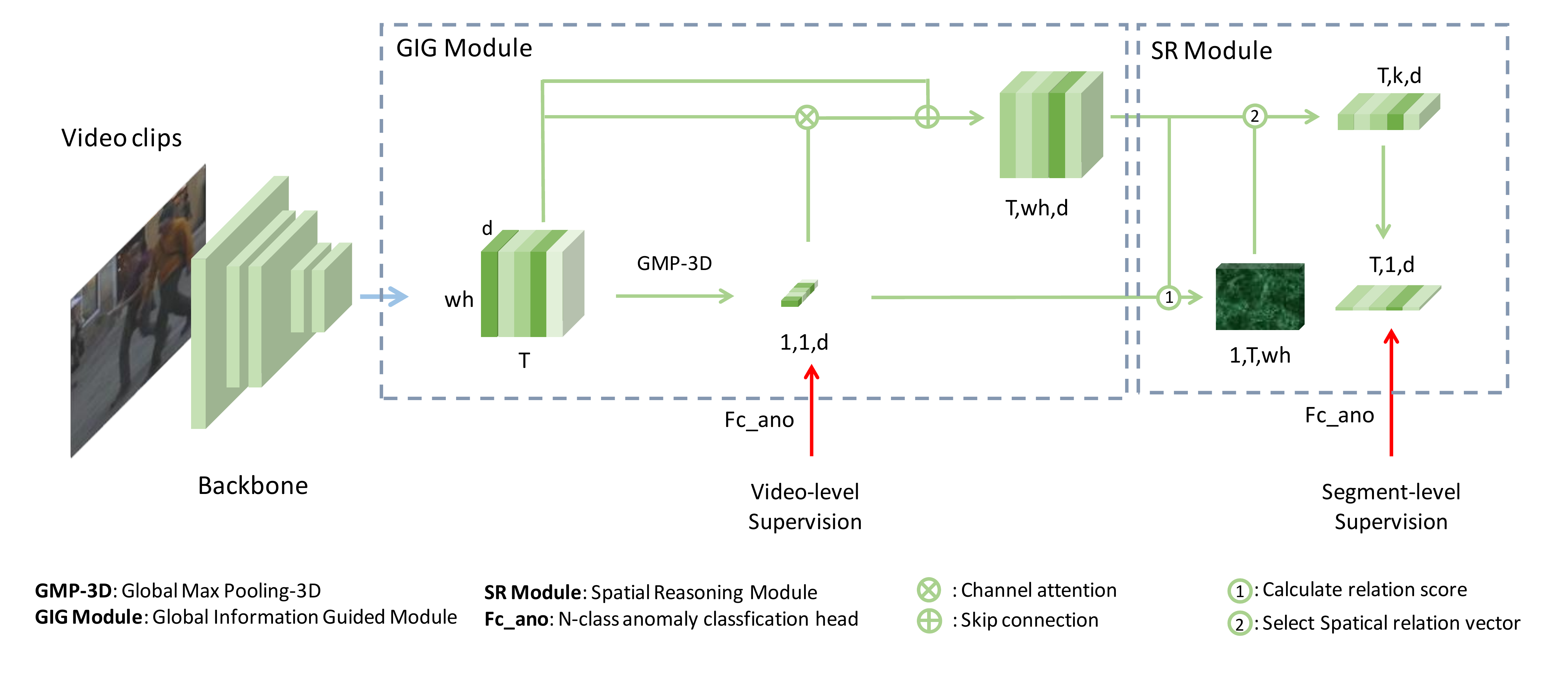}
	\caption{Overall pipeline of our Global Information Guided (GIG) anomaly detection framework. The whole architecture consists a backbone model and the designed video anomaly detection model. In the VAD model, we propose a GIG module that generates the global pattern vector for distinguishing whether an anomaly occurs in a video. The pattern vector is further used to enhance the features by an attention mechanism. In addition, we introduce a Spatial Reasoning (SR) module for selecting the most anomaly-relevant vector in spatial domain for later anomaly classification. The dimension of extracted feature maps from the backbone is $\mathbb{R} ^ {T \times w \times h \times d}$. $T$, the temporal dimension, represents the amount of video segments. $w,h$ are spatial dimensions, we stack them into one dimension for visualization and $d$ denotes the channels of feature maps. }
	\label{Pipeline}
\end{figure*}

In this paper, an end-to-end anomaly detection framework is designed to facilitate anomaly detection in videos with a high confidence and efficiency. 
We first utilize a backbone model to extract features from video-clips, upon which we build a Global Information Guided (GIG) module to localize anomalies in temporal domain. 
The whole network architecture is depicted in Figure~\ref{Pipeline}. 
In the designed GIG module, we first leverage the video-level label to capture the global pattern cue in a video sequence. Then we enhance the spatio-temporal feature representation with the global information by channel-wise attention mechanism.

In addition, spatial information is largely neglected in previous approaches~\cite{sultani2018real,zhong2019graph,zhu2019motion}. 
They extracted the semantic features after the \emph{global\_pooling} operation from pre-trained model. 
In this way, the spatial resolution is directly reduced to $1 \times 1$, hence the appearance cue and relationship among objects are restrained largely. 
While these informations are of great important for detecting and distinguishing anomalies. 
For example, $Robbery$ describes the action between people, $Stealing$ usually happens between people and property, and if a gun is holding in one's hand, it is easy to witness a $Shooting$ event.
Inspired by this, besides the motion semantics extracted by action classification model, we further investigate the spatial cues by exploring the relevance between spatial feature vectors and global pattern cue vector in a $Spatial$ $Reasoning$ (SR) Module.
Then we select the most relevant spatial vectors for later anomaly classification and localization. 

To tackle the weak supervision VAD, with only the video-level annotations available, we propose a novel Video-Segment objection function, namely $VS$ $Loss$, for utilization of video-level annotations and expanding it to segment-level supervision.
The whole function contains two components, which are $Video-level$ $Supervision$ and $Segment-level$ $Supervision$. $Video-level$ $Supervision$ is for discovering the global pattern information of each video sequence, which can also serve as an intermediate supervision. $Segment-level$ $Supervision$ is designed to choose the segments with maximum probabilities of certain anomaly and strengthen the corresponding probabilities.

The main contributions of this paper are as follows:
\begin{enumerate}
	\item We design an end-to-end video anomaly detection (VAD) framework to classify and localize anomalies with a high convenience and efficiency. Our approach achieves good results on CityScene challenge, which verifies the effectiveness of our method.
	\item We propose a global information guided anomaly detector that explores the global pattern cue first and leverages the spatial information for VAD.
	\item We introduce a novel Video-Segment objection function ($VS$ $Loss$). This function enables the video-level and segment-level supervision at the same time for a better mining of the global information.
\end{enumerate}
\fancyhead{}
\section{Related Work}
In this section, we briefly review the techniques of video anomaly detection under weak supervision.
Recently there are researches employing normal and abnormal video data along with video-level annotations, for model building \cite{sultani2018real,he2018anomaly,zhu2019motion,zhong2019graph}.
Among them, Multiple Instance Learning (MIL) is introduced for pattern modeling~\cite{sultani2018real,he2018anomaly,zhu2019motion}.
Sultani et al. \cite{sultani2018real} considered anomaly detection as a MIL problem with a novel ranking loss function. Later, by extending it, Zhu et al. \cite{zhu2019motion} introduced the attention mechanism for better localizing anomalies. Due to the absence of anomaly positions in training phase, these two methods cannot predict anomaly frames well. For this, Zhong et al. \cite{zhong2019graph} attempted to construct supervised signals of anomaly positions through iteratively refining them. 
Although great progress has been witnessed in the domain, the spatial information is neglected in previous VAD methods to a large extent. Different from existing methods, we propose to mine the global pattern cue within each video, with the benefit of video-level annotation. Then the global information is used to measure the relevance of spatial feature vectors. The most related vectors are cooperated for anomaly classification and localization.

\section{The Proposed GIG Anomaly Detection Framework}
\subsection{Overview}
The whole pipeline of our GIG model is shown in Figure~\ref{Pipeline}. Video clips are taken as the input of the backbone network to extract spatio-temporal feature maps. Given the training data $\{\textbf{V}\}_{i=1}^N$ of $N$ video sequences, we divide each video into $T$ segments. Each anomalous video is equipped with a class label set, containing $n_s$ unique anomaly instances represented as $\textbf{y} = \{y^k\}^{n_c}_{k=1}$, where $y^k \in \mathcal{A}$, the set of all anomaly classes with a total class number of $C$. Within each segment, we sample $n_s$ video clips with equal time interval. The sampled clips are taken to represent the segment. Upon the backbone model, we design a Global Information Guided (GIG) model that leverages the video-level labels for anomaly detection.
\subsection{Global Information Guided Module}
Here, we explain the details of our Global Information Guided model for video anomaly detection. In this model, we propose to mine the Global Pattern Cue (GPC) within each video sequence and utilize the global information to enhance the anomaly representation. 
Given the extracted high-level semantic feature maps $\textbf{X} = \{x^i\}_{i=1}^T$ of all segments, with resolution $w,h,d$ of $x$, we can obtain the global pattern cue by spatio-temporal dimension reduction operation. We apply the simple and effective $Global Max Pooling-3D$ function here as:
\begin{align}
\textbf{g} = \psi (\textbf{X}),
\end{align}
where $\psi$ denotes the dimension reduction operation, $\textbf{g}$ is the vector of the GPC with resolution of $\mathbb{R} ^ {1 \times 1 \times 1 \times d}$.

Inspired by the great progress in channel attention~\cite{li2019selective,hu2018squeeze}, we adopt a simple channel-wise attention operation to imply the GPC vector $\textbf{g}$ as a guidance for representation enhancement, formally:
\begin{align}
\widehat{\textbf{X}} = \sigma(\textbf{g}) \odot \textbf{X} +\textbf{X},
\end{align}
here $\widehat{\textbf{X}}$ represents the enhanced feature maps, $\sigma$ is a $sigmoid$ function for normalizing the weight scalars to $[0,1]$ and $\odot$ denotes the channel-wise multiplication operation. For preserving the original information, we leverage the skip connection to generate the final representation.

Further we propose to take the video-level annotation $\textbf{a}$ as the supervision upon the GPC, which can also be viewed as an intermediate supervision. At First, we employ a fully connected function $\phi_1$ as the anomaly classification head to measure the anomaly status, using the GPC vector as input and outputting scores of $1+C$ classes (one normal class and $C$ anomaly class). Here we select to analyze the anomalous status of the GPC vector in a coarse level by detecting whether any anomaly instance exists in the video, rather than detecting which the anomaly class is, if exists.
In detail, we take the maximum score among $C$ anomaly classes of the GPC vector as the overall anomaly score $S_{\textbf{g}}^*$: 
\begin{align}
S_{\textbf{g}}^*=max_{i=1}^C(\sigma(\phi_1(\textbf{g}))_i),
\end{align}
where $\sigma(\phi(\textbf{g}))_i$ denotes the anomaly score of $i$th class. After the maximum operation, the global anomaly score becomes a two-channel vector, indicating the normal and abnormal probability respectively. Then we apply a $Video-level$ $Supervision$ with $Binary$ $Cross$ $Entropy$ loss as:
\begin{align}
\ell_{\textbf{g}}^* = -(\textbf{y}^* \log S_{\textbf{g}}^*+(1-\textbf{y}^*)\log (1-S_{\textbf{g}}^*)).
\end{align}
Here, $\textbf{y}^*$ is set to 1 if the video contains any anomaly instance, otherwise 0.
\subsection{Spatial Reasoning Module}
After mining the global information, we further analyze the relationship between each spatio-temporal vector and the GPC vector to retrieve spatial cue vital for anomaly classification and localization. Previous methods~\cite{sultani2018real,zhong2019graph,zhu2019motion} usually apply a \emph{global\_pooling} operation to integrate the spatial information, however in this way, the object appearance cues and the interaction between objects are largely neglected, which is of great importance for distinguishing the anomalies. For example, the instances of $Robbery$, $Stealing$ and $Shooting$ are prone to false detection. $Robbery$ describes the event between at least two people, $Stealing$ usually happens between people and property, and if a gun is holding in one's hand, it is easy to witness a $Shooting$ event.

In the designed SRM, we take the GPC vector $\textbf{g}$ and the enhanced spatio-temporal feature maps $\{\widehat{\textbf{x}}\}_{i=1}^T \in\widehat{\textbf{X}}$ as inputs. At the first step, we measure the spatial key by calculating the relation score between $\textbf{g}$ and $\widehat{\textbf{X}}$, formally:
\begin{align}
r_{i,j}=R(\textbf{g},\widehat{\textbf{x}}_{i,j}),
\end{align}
with $i,j$ denoting the row and column index, $r$ as the relation score, and $R$ as the Reasoning function. Here we choose the $cos$ $simliarity$ as the relation measurement metric.
At the second step, we choose the $k$ spatial vectors with the top-$k$ relation scores and further reduce the spatial dimension to 1 for obtaining the spatial pattern vector $\widehat{\textbf{x}}_{s}$ of each segment, by aggregating the spatial vectors corresponding as:
\begin{align}
\widehat{\textbf{x}}_{s}=\frac{1}{k}\sum_{i\in[1,w],j\in [1,h]}top\_k(\widehat{\textbf{x}}_{i,j},r_{i,j}).
\end{align}
Then we apply an anomaly classification head $\phi_2$ for generating the pattern vectors of each segment as:
\begin{align}
S_{\textbf{s}}=\sigma(\phi_2(\widehat{\textbf{x}}_{s})).
\end{align}
Once we obtain the anomaly scores $\{S_{\textbf{s}}\}_{i=1}^T$ of $T$ segments of a video, we need to summarize the results from segment-level to video-level.
In detail, we pick up the top $p$ maximum segment-level scores and average them to make a segment-consensus score as:
\begin{align}
S_{\textbf{s}}^T=\frac{1}{p} \sum_{i=1}^T top\_p(S_{\textbf{s}}^i).
\end{align}
We apply the $Segment-level$ $Supervision$ on the segment-consensus score.
At first, we enable a supervision on the overall anomalous status as the GPC vector, formally:
\begin{align}
\label{f9}
{S_{\textbf{s}}^*}&=max_{i=1}^C(S_{\textbf{s}}^T)_i,\\
\ell_{\textbf{s}}^* &= -(\textbf{y}^* \log S_{\textbf{s}}^*+(1-\textbf{y}^*)\log (1-S_{\textbf{s}}^*)),
\end{align}
Similar as the GPC vector, ${S_{\textbf{s}}^*}$ is the binary overall anomaly score and $\ell_{\textbf{s}}^*$ is the corresponding loss term.
We further distinguish the anomaly instances by supervising the segment-consensus anomaly score under multi-class anomaly detection function as:
\begin{align}
\ell_{\textbf{s}} &= -(\textbf{y} \log S_{\textbf{s}}^T+(1-\textbf{y})\log (1-S_{\textbf{s}}^T)),
\end{align}
here, \textbf{y} is the multi-hot anomaly label and $\ell_{\textbf{s}}$ is the anomaly classification loss.
The $Segment-level$ $Supervision$ and $Video-level$ $Supervision$ make up our Video-Segment loss $\ell_{\textbf{vs}}$.
\begin{align}
\label{vs}
\ell_{\textbf{vs}} &= \ell_{\textbf{s}} + \lambda_1 \ell_{\textbf{s}}^* +\lambda_2 \ell_{\textbf{g}}^*,
\end{align}
with $\lambda_1$, $\lambda_2$ representing the balance weights for various terms.
Further more, since the anomaly mostly occurs for a short period of time in real life, we also add a sparse constraint as in~\cite{zhu2019motion}:
\begin{align}
\ell_{\textbf{sparse}} = \sum_{i=1}^T S_{\textbf{s}}^i,
\end{align}
Finally, we collect all the loss terms to get the overall supervision signal $\ell$, with $\lambda_3$ being the weight for the sparse term and $\ell_{\textbf{vs}}$ from Eqn.~\ref{vs}. Formally:
\begin{align}
\ell &= \ell_{\textbf{vs}} + \lambda_3\ell_{\textbf{sparse}} \nonumber\\
&= \ell_{\textbf{s}} + \lambda_1 \ell_{\textbf{s}}^* +\lambda_2 \ell_{\textbf{g}}^*+\lambda_3 \ell_{\textbf{sparse}}.
\end{align}

\section{Experiments}
\subsection{Dataset}
We conduct method on the CityScene challenge~\cite{CityScene}. CitySCENE dataset consists of videos which cover 12 real-world anomalies related to public safety and city management, including $Accident$, $Carrying$, $Crowd$, $Explosion$, $Fighting$, $Graffiti$, $Robbery$, $Shooting$, $Smoking$, $Stealing$, $Sweeping$, and $WalkingDog$. The training set consists of 758 normal and 1319 anomalous videos. The videos in training set are trimmed manually and the labels are at video-level. Videos in testing set are untrimmed, with a total amount of 188. In addition, anomaly mostly occurs for a short period of time in testing videos.
\subsection{Implementation Details}
We adopt the Temporal Pyramid Network (TPN)~\cite{yang2020tpn}, which is a state-of-the-art action classification network, as our backbone model. We use the resnet50 version of TPN and extract feature maps after the TPN module in~\cite{yang2020tpn} with the resolution of $\mathbb{R} ^ {T \times 7 \times 7 \times 2048}$. We load the model pre-trained on kinetics400 as the default weights. 
Empirically, we set the number of video segments $T$, to 8. In each segment, we randomly sample 6 video clips with a time interval of 5 frames.
During the training phase, we set the batch size as 8.
We employ Adagrad~\cite{duchi2011adaptive} optimizer with the learning rate of $0.001$ for $100$ epoch.
The augmentation of horizontal flip and a dropout of 0.5 are adopted to reduce overfitting.
The hyper-parameters $\lambda_1$, $\lambda_2$ and $\lambda_3$ are set to $1,0.5,0.1$ respectively.
It is worth mentioning that we do not use any extra video data in the challenge.
At testing time, we predict an anomaly score for every $6$ video frames with a stride of $3$. After getting the rough anomaly in each video, we apply a $gaussian$ $filter$ with sigma $2$, order $0$ for smoothness.
The experiments are conducted on 4 2080Ti Nvidia GPU.
\begin{table}[!tbp]
	\centering
	\caption{General Anomaly Detection Rank List}
	\scalebox{0.9}{
		\begin{tabular}{ccccc}
			\toprule
			Rank&Method &AUC(\%)&&\\
			\midrule
			1&BigFish&89.20\\
			2&MonIIT&87.94\\
			3&DeepBlueAI&86.85\\
			4&SYSU-BAIDU&86.52\\
			5&UHV&85.37\\
			\textbf{6}&\textbf{Ours}&\textbf{84.09}\\
			7&ActionLab &70.92\\
			8&Orange-Control&31.65\\
			\bottomrule
	\end{tabular}}	
	\label{t1}
\end{table}
\begin{table}[!htbp]
	\centering
	\caption{Specific Anomaly Detection Rank List}
	\scalebox{0.9}{
		\begin{tabular}{ccc}
			\toprule
			Rank&Method &MF1(\%)\\
			\midrule
			1&SYSU-BAIDU&66.41\\
			2&BigFish&62.11\\
			\textbf{3}&\textbf{Ours}&\textbf{52.33}\\
			4&MonIIT&45.52\\
			5&Orange-Control&40.42\\
			6&UHV&38.36\\
			7&ActionLab &22.22\\
			\bottomrule
	\end{tabular}}	
	\label{t2}
\end{table}
\subsection{Evaluation metrics}
This CityScene challenge has two tasks: (1) general anomaly detection, which treats all anomalies in one group and all normal events in the other group; and (2) specific anomaly detection, which recognizes each of the anomalous activities.
For task 1, frame-based receiver operating characteristic (ROC) curve and corresponding area under the curve (AUC) to evaluate the performance of the method. An ROC space is defined by FPR and TPR as x and y axes, respectively, which depicts relative trade-offs between true positive (benefits) and false positive (costs).
For task 2, frame-based F1-score is selected as the evaluation metric to evaluate the performance. For each predefined class $c$, the precision $precision^c = \frac{{TP}^c}{{TP}^c+{FP}^c}$ and recall $recall^c = \frac{{TP}^c}{{TP}^c+{FN}^c}$ from the results are calculated to get the F1-score $F^c_1 = 2*\frac{{precision}^c*{recall}^c}{{precision}^c+{recall}^c}$. (the frames not included in the results will be labeled as negative.) The mean of class-wise F1-score (MF1) is chosen as the overall metric of task2.
\subsection{Experimental Results}
We have participated in both task1 and task2 in CityScene challenge. The results are listed in Table~\ref{t1} and Table~\ref{t2} respectively.
As to task 1, we obtain the frame-level anomaly score as in Formulation~\ref{f9} from the spatial vectors. The AUC of our method reaches up to 84.09\%, superior than ActionLAb (70.92\%) and Orange-Control (31.65\%). Although the AUC performance of our method is  lower than the best approach, the gap is small. For task 2, our method wins the 3rd place in the challenge with MF1 of 52.33\%, which exceeds the 4th method with a margin of 7\%. In addition, our GIG model can run at almost $100$ fps on a 2080Ti GPU. The fast speed, as well as the state-of-the-art performance shows the superiority of our GIG model.

\section{Conclusion}
In this work, we propose an end-to-end Global Information Guided anomaly detection framework for anomaly classification and localization. It is the first end-to-end anomaly detection model under weak annotations, as far as we know. We leverage the video-level labels to mine the global pattern cue of the entire video sequences as first and enhance the learned representation with the global pattern information by channel-wise attention. Then we retrieve the spatial cue in each video segment by filtering the spatial vectors with the most relevance with the global pattern cue to obtain the anomaly vectors of the segments. The experimental results demonstrate the effectiveness of our method with a high speed and detection accuracy. 
\section{Acknowledge}
This work was supported by the National Natural Science Foundation of China (Grants Nos.61772276).
\bibliographystyle{ACM-Reference-Format}
\bibliography{sample-base}
	
\end{document}